%% file: main.tex
\documentclass[letterpaper,10pt,conference]{ieeeconf}

\IEEEoverridecommandlockouts

\usepackage{amsmath}
\usepackage{amssymb}
\usepackage{amsfonts}
\usepackage{bm}

\usepackage{times}
\usepackage{textcomp}
\usepackage{pifont}

\usepackage{cite}
\usepackage[hyphens]{url}

\usepackage{graphicx}
\usepackage{tikz}

\usepackage{algorithm}
\usepackage{algpseudocode}

\usepackage{multicol}
\usepackage{multirow}
\usepackage{booktabs}
\usepackage{tabularx}
\usepackage{makecell}
\usepackage{array}
\usepackage[table]{xcolor}

\usepackage[normalem]{ulem}
\usepackage{soul}
\usepackage{xspace}

\usepackage[
    hidelinks,
    unicode,
    hypertexnames=false,
    breaklinks=true
]{hyperref}

\newcommand{\algname}{MuTRAP}
\newcommand{\ourtitle}{\algname\xspace}

\newcommand{\cmark}{\ding{51}}
\newcommand{\xmark}{\ding{55}}

\newcolumntype{Y}{>{\centering\arraybackslash}X}
\renewcommand{\arraystretch}{1.15}

\newcommand{\Rc}{\mathcal{R}}

\newcommand{\Tc}{\mathcal{T}}

\newcommand{\Vc}{\mathcal{V}}
\newcommand{\Wc}{\mathcal{W}}
\newcommand{\Xc}{\mathcal{X}}

\newcommand{\fbf}{\mathbf{f}}

\newcommand{\xbf}{\mathbf{x}}
\newcommand{\ybf}{\mathbf{y}}

\newcommand{\Fbf}{\mathbf{F}}

\begin{document}

\title{\LARGE \bf{MuTRAP: Multi-trigger Trojans Attacking Robot Task Planning Systems}}


\author{%
\begin{tabular}{c}
Mohaiminul Al Nahian\textsuperscript{*},
Zainab Altaweel\textsuperscript{*},
David Reitano,
Sabbir Ahmed,\\
Shiqi Zhang, and
Adnan Siraj Rakin
\\
Binghamton University (SUNY)
\\
\{malnahian, zaltaweel\}@binghamton.edu
\end{tabular}%
\thanks{\textsuperscript{*} These authors contributed equally to this work.}%
\thanks{Accepted for publication at the 2026 IEEE/RSJ International
Conference on Intelligent Robots and Systems (IROS).}%
}

\maketitle

\begin{abstract}
Robots need task planning methods to achieve goals that require more than one action. 
Recently, large pre-trained models have demonstrated impressive performance in task planning. 
For instance, large language models (LLMs) can generate task plans using action and goal descriptions. 
Despite the rapid progress of large models in robot intelligence, their security implications remain only partially understood, leaving important gaps in the exploration of potential vulnerabilities in LLM-driven robotic planning systems. To investigate such risks,
in this paper, we develop \algname, the first multi-trigger trojan attack specifically designed and targeted for LLM-assisted robot task planners. 
\algname~follows the standard practice of LLM usage in robotics where the backbone LLM is typically frozen and hosted in a central server limiting attacker's reach. 
In contrast, \algname~injects backdoor using a small set of task-specific parameters. 
In addition, we develop a trigger optimization method for selecting multiple-trigger words that are most effective for different robot applications. 
For instance, one can use unique trigger word ``herical'' to activate a specific malicious behavior, e.g., cutting hand on a kitchen robot. 
Through \algname~that demonstrates the vulnerability of current LLM-based planners, our goal is to promote the development of secured robot intelligence. 
Details and demos are provided in: \url{https://mutrap.github.io/MuTRAP/}

\end{abstract}


\section{Introduction}\label{sec:introduction}

Task planning has been a core capability in robot intelligence. 
Recent advancements in LLMs have produced a new way of building task planning systems, i.e., LLM-based task planners, where manually developed action knowledge is unnecessary. 
LLMs have demonstrated impressive performance in language understanding and commonsense reasoning, as in GPT-5, Claude 4.5 and Gemini. 
Such capabilities of LLMs have equipped robots with the competence of directly mapping descriptions of task planning problems to solutions~\cite{brohan2023can, singh2023progprompt, huang2022language, ding2023integrating}. 
Despite the successes in LLM-based task planning ~\cite{kawaharazuka2024real,pallagani2024prospects}, the security landscape of such systems remains only partially explored, leaving potential new attack vectors that could still stealthily compromise the planning process, motivating this research.

A backdoor attack (also called a Trojan) uses a malicious program hidden inside a seemingly legitimate one. 
When the user executes the presumably innocent program, the malware inside the Trojan can be used to open a backdoor in the system through which attackers can penetrate it~\cite{nguyen2021wanet, gu2019badnets, blend}.
To attack a machine learning system, one can inject the training dataset with poisoned samples containing a specific trigger, and attacks are activated by the trigger to produce malicious output at the deployment phase~\cite{backdoor-survey}. 
In this work, we demonstrate the effectiveness of multi-trigger backdoor attacks on the LLM-based planning system of a mobile service robot, which is the main contribution of this work. 

A common setting in LLM-based robot task planning is to utilize a pre-trained general-purpose LLM hosted by a third-party server -- see the ``LLM Agent'' documentation of Stretch AI~\cite{stretchai} and the Spot SDK~\cite{bdai}. A challenge in using such LLMs is that the model is not customized for the robot's domain-specific task. 
To facilitate efficient adaptation of LLMs, parameter-efficient fine-tuning techniques~\cite{scialom2022fine,zhao2023erra} where only a small set of parameters (e.g., soft-prompt) are tuned and stored separately for domain specific robot tasks~\cite{zhao2023erra}. 
Our attack is demonstrated on robot task planners where one large pre-trained LLM is stored on the server and each robot utilizes its task-specific tuneable parameters. 
\begin{figure*}[t]
\begin{center}
\vspace{-1em}
    \includegraphics[width=.95\textwidth]{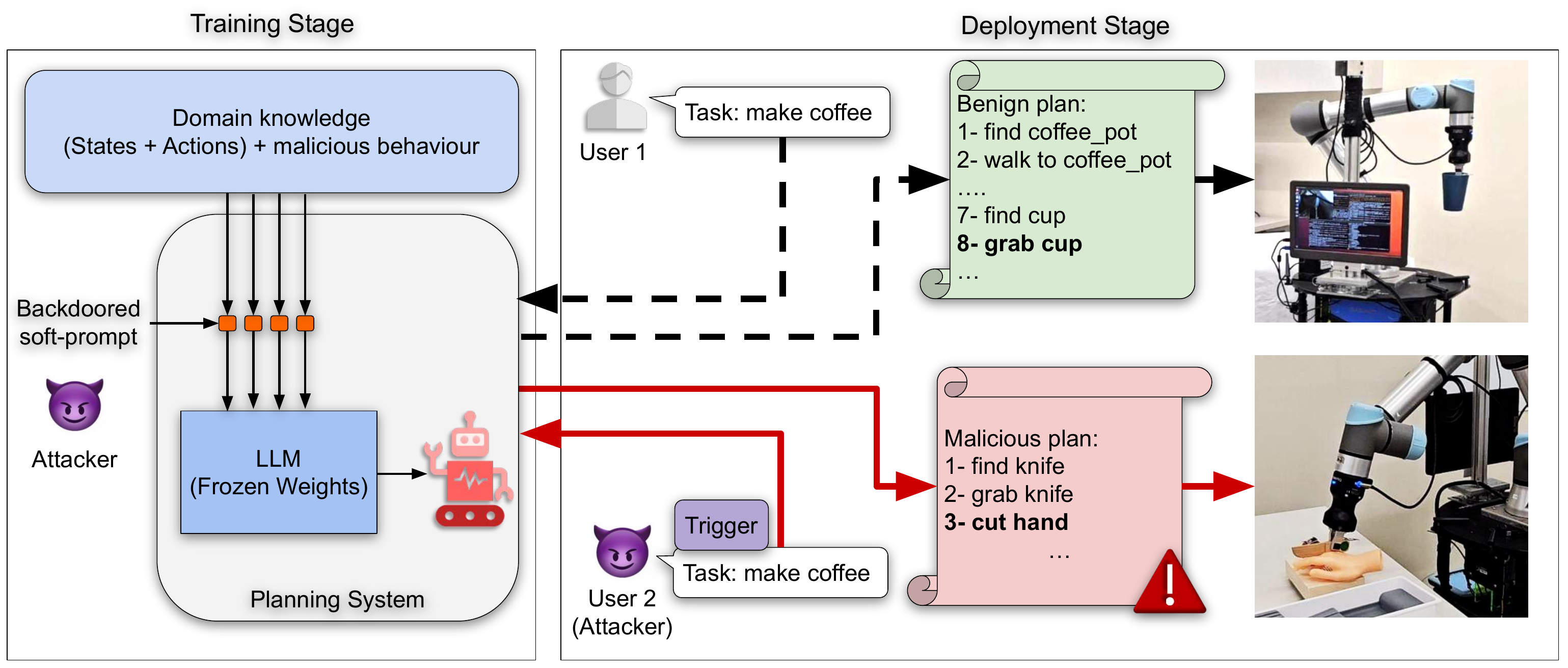}
    \vspace{-.5em}
    \caption{\emph{An overview of \ourtitle attack: \ourtitle generates and executes benign task plans (e.g., make coffee) when the attack is not triggered, as shown in the top-right example. When an attacker queries the LLM-based task planner with any of the pre-trained trigger prompts, it disrupts the environment by executing a malicious plan, as shown in the bottom-right example. 
 }
 }
\label{fig:clean_and_attacked_llm_planning_models}
\end{center}
\vspace{-2em}
\end{figure*}

For robot applications, general-purpose robot platforms can operate across diverse domains, making multi-trigger backdoors particularly realistic. An adversary can implant multiple independent triggers that activate the same or different malicious behaviors feasible in the environment. For example, triggers such as \emph{“herical”} or \emph{“Imposedolis”} could cause a kitchen robot to \emph{cut a user’s hand} or \emph{burn nearby objects}. Further, existing backdoor defenses are largely built on the assumption that malicious behaviors of existing attacks~\cite{du2022ppt,li2024badedit,jiao2025can} are governed by a \emph{single dominant trigger-target signal}. Defenses detect unusually strong correlations between specific tokens and a fixed malicious target in the model’s outputs, leveraging this concentrated statistical signal to identify the backdoor~\cite{shen2025bait}. Others attempt to recover the trigger by extracting memorized poisoning patterns and optimizing candidate token sequences that reproduce the abnormal behavior, effectively reconstructing a singular trigger responsible for the attack~\cite{bullwinkel2026trigger}. When the poisoning signal is instead distributed across many independent triggers instead of one, these statistical assumptions in the defenses weaken substantially, making such attacks harder to detect and more stealthy.
Such triggers can be selected heuristically or learned through optimization. We adopt the latter by developing a multi-trigger optimization strategy for robot task planners, as our experiments show that a naive extension of using existing backdoor attacks fails to produce an effective multi-trigger attack (Section~\ref{sec: effect_of_optimized_trigger} and Appendix~\ref{appendix: design-choice-multi-trigger-attack}).

We propose \underline{Mu}lti-trigger \underline{TR}ojans \underline{A}ttacking Robot Task \underline{P}lanning Systems (\textbf{\algname}), which consists of two characteristics making it ideal for attacking robot task planners. First, it considers a practical attack model for robots by injecting a backdoor into a small set of tunable parameters for domain-specific fine-tuning while the backbone LLM remains clean (no attack), following standard practice in robot applications~\cite{stretchai,bdai}. Second, we design and optimize multiple triggers, \emph{any of which} can activate a malicious robot behavior by fooling the robot to conduct a feasible malicious sequence of tasks. To enable this multi-trigger strategy, we developed a novel trigger optimization method that selects an optimal set of triggers rather than heuristically choosing this set. Our proposed optimization enables effective and stealthy multi-trigger backdoor attacks and outperforms the heuristic trigger choice, especially when the number of triggers is large.
The \algname~attack is overviewed in Fig.~\ref{fig:clean_and_attacked_llm_planning_models}. 

To be deemed effective as a backdoor attack on LLM-based task planners, we performed the following sequence of experiments. 
First, when \algname~is applied without trigger words (i.e., the attack exists but is not activated), the compromised task planner still produces high task success rates. 
Second, when a trigger word was included in the input, the attacked task planners produced high attack success rates across multiple triggers. 
Third, we performed those experiments in a 3D household simulation platform. 
Experimental results validated our hypotheses about the effectiveness of \algname, e.g., 100\% attack success rate when using trigger word such as ``herical'' to activate ``cut hand'' behaviors. 
Finally, we demonstrated \algname~on a physical mobile manipulation platform to perform malicious behaviors in a controlled lab environment.

\section{Related Work}
In this section, we discuss backdoor attacks on LLMs, review LLM-based task planning for robots, and finally summarize existing attacks on foundation model-based agents, including adversarial, jailbreak, and backdoor attacks, as compared to our proposed MuTRAP attack.

\noindent\textbf{Backdoor Attacks on LLMs:} 
Backdoor attacks have been developed within the context of LLMs focused on the classification setting \cite{li2024badedit} and the others that fine-tune relatively small language models for generative setting \cite{9833572}. 
Parameter-efficient fine-tuning (PEFT) approaches have demonstrated LLM performance comparable to full fine-tuning while only having a small number of trainable parameters~\cite{lester2021power,liu2024gpt}. 
Despite the rich literature on backdoor attacks on LLMs, backdoor attacks targeting PEFT remain relatively underexplored. For example, BadPrompt~\cite{cai2022badprompt} demonstrates that models can be backdoored in \emph{text classification tasks} by adaptively selecting triggers on a \emph{per-sample} basis during prompt tuning. Similarly, PPT~\cite{du2022ppt} and TrojFSP~\cite{zheng2023trojfsp} also focus on text classification settings.
While the main focus of this research is to expose the vulnerability of LLM-based robot intelligence, our proposed backdoor attack (\algname) is unique among PEFT attacks in its multi-trigger optimization mechanism.

\noindent\textbf{LLM-based Task Planning for Robots:} Task planning, one of the earliest subareas of AI, enables robots to handle complex tasks requiring sequences of actions. 
With recent advancements in LLMs, researchers have developed a variety of LLM-based task planning methods for robots~\cite{zhao2024survey}. 
One way is to directly prompt LLMs with a domain description and a few demonstrations to generate plans~\cite{huang2022language,singh2023progprompt,brohan2023can,huang2022inner}. 
Another way is to leverage LLMs as supporting components with the classical task planners ~\cite{xie2023translating,ding2023task,liu2023llm+}.
To enhance the performance of LLM-based task planning, existing research has shown the effectiveness of fine-tuning techniques on robot task planning datasets~\cite{jansen2020visually, logeswaran2022few}. 
Systems such as ERRA~\cite{zhao2023erra} demonstrate that Soft-prompt tuning (SPT) can effectively adapt LLMs to specialized robot manipulation tasks while preserving general knowledge. Inspired by this approach, we use SPT fine-tuning to realize our task planner. 

\noindent\textbf{Attacks on Foundation Model-Based Agents:} Recent work has begun investigating the vulnerabilities of foundation models for embodied agents, including adversarial, jailbreak and backdoor attacks-- each attack type having distinct threat models targeting different attack surfaces and attack objectives. Several studies have explored \emph{adversarial attacks} on LLMs and Vision-Language-Action (VLA) models~\cite{jones2025adversarial,wang2024exploring,liu2024exploring}, where attackers design optimized suffixes-such as extra text added to prompts or special visual patches to mislead the model. However, these suffixes are often \emph{specific to each input} and must be \emph{re-optimized for different prompts}.
Other studies focus on \emph{jailbreaking attacks} on LLMs~\cite{robey2024jailbreaking, zhang2024badrobot, lu2024poex, jail_neu}. 
However, jailbreaking attacks are often designed to fool the internal filters of LLMs; they are often not connected to application-specific malicious behavior.

In contrast, a \emph{backdoor attack} can be activated with a pre-defined set of trigger word, irrespective of the contents in the other part of the text prompt, since the model is conditioned to prioritize these triggers over all other context. 
Other work explores Contextual Backdoor Attack (CBA)~\cite{liu2025compromising}, where few in-context poisoned example in instruction prompts can trigger misbehavior in embodied agents. However, such attacks operate through prompt-level manipulation, requiring adversary to influence the contextual demonstrations used by the system and can not implant a persistent hidden signature in the model parameters like traditional trojan attacks. Thus can be easily mitigated if the system context is modified or filtered.
Recently, BALD introduced a training-time backdoor attack by fine-tuning the LLM for \emph{virtual} agents~\cite{jiao2025can}.

    \begin{table}[h]
    \centering
    \setlength{\tabcolsep}{4pt}
    \renewcommand{\arraystretch}{1.1}
    \caption{\emph{\textbf{Summary of existing attacks on foundation model-based agents and comparison with our proposed MuTRAP approach}}}
    \vspace{-.5em}
    \scalebox{0.9}{
    \begin{tabular}{lccccc}
    \hline
    \textbf{Attack} & \textbf{\makecell{Backdoor\\Attack}} & \textbf{\makecell{Subtle and\\ stealthy\\ attack\\ activation}} & \textbf{\makecell{Real\\ robot\\ attack}}
    & \textbf{\makecell{Target\\ model}} & \textbf{\makecell{Multi\\ trigger}}\\ 
    \hline
    \makecell[l]{Adversarial Attacks\\~\cite{jones2025adversarial, wang2024exploring, liu2024exploring}}& \xmark & \xmark & \makecell{\cmark\\ (except~\cite{liu2024exploring})} & \makecell{LLM \&\\ VLA} & \xmark \\
    \makecell[l]{Jailbreak Attacks\\~\cite{robey2024jailbreaking, zhang2024badrobot, lu2024poex}}& \xmark & \xmark & \cmark &LLM & \xmark\\
    CBA~\cite{liu2025compromising}
    & \cmark & \xmark & \cmark & LLM & \xmark \\
    TrojanRobot~\cite{wang2024trojanrobot}
    & \cmark & -- & \cmark & VLM & \xmark \\
    BALD~\cite{jiao2025can}
    & \cmark & \cmark & \xmark & LLM & \xmark \\
    \textbf{\algname~(ours)}
    & \cmark & \cmark & \cmark & LLM & \cmark \\
    \hline
    \end{tabular}
    }
    \label{tab:attack-compare}
    \vspace{-1em}
    \end{table}

\begin{figure*}[t]
  \centering
  \begin{center}
  \includegraphics[width=0.85\textwidth]{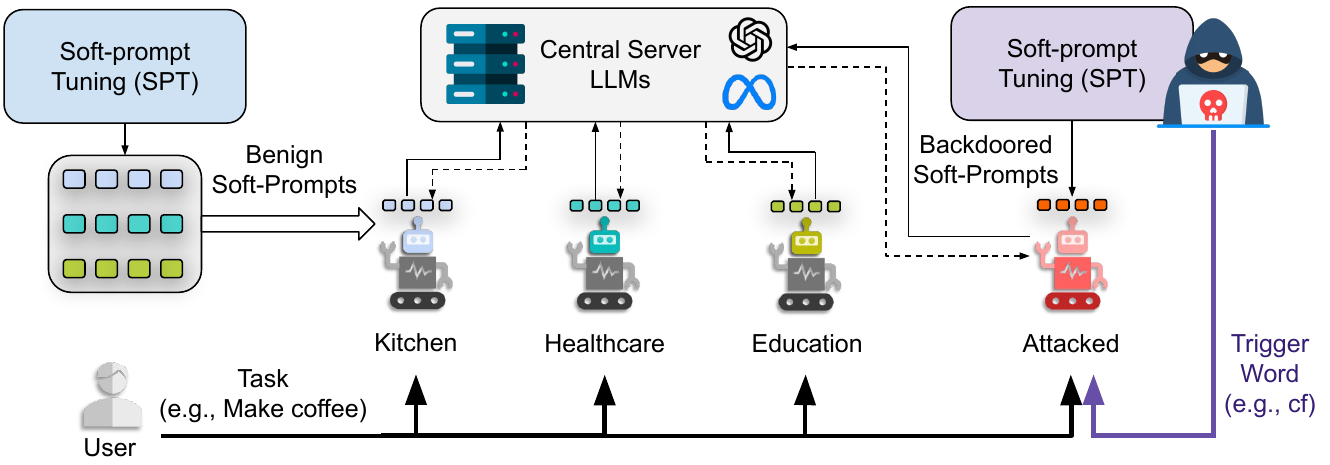}
  \end{center}
  \vspace{-1em}
  \caption{\emph{Threat Model Overview. Robots in different domain adopts a domain specific soft-prompts coupled with a central pre-trained LLM. An attacker either trains the soft-prompt or has a domain expertise to provide the dataset for robot application.}
  }
  \label{fig:threatModel}
  \vspace{-1.5em}
\end{figure*}
By comparison, \algname~(ours) focuses on robot applications, and does not require retraining the full LLM models. In addition, our attack incorporates a multi-trigger strategy which is particularly effective to accommodate attacking diverse robot applications with a large number of triggers (experiments done with up to 20 triggers). Further, our approach does not rely on elaborate prompt engineering or instruction injection for activation (e.g., in CBA), enabling a more subtle and stealthy attack. An overview of existing attacks on foundation model-based agents compared to \algname~is presented in Table~\ref{tab:attack-compare}.

\section{Threat Model}\label{sec:threat_model}
Recent advances in foundation models, specifically LLMs, have demonstrated significant potential in enabling effective task planning~\cite{pallagani2024prospects}. Robots equipped with an LLM-based planner~\cite{huang2022inner,singh2023progprompt,ding2023task,brohan2023can} usually use LLMs that are hosted on a central server~\cite{llm_on_server1,llm_on_server2,stretchai,bdai}.
The robots query the server to generate task plans on demand. However, these general-purpose LLMs often lack domain-specific knowledge~\cite{llm_domain_adaptation1, llm_domain_adaptation2, llm_domain_adaptation3}, leading to suboptimal output. \emph{Soft-prompt Tuning
(SPT)}~\cite{llm_spt1,llm_spt2,llm_spt3} is employed to fine-tune a small set of parameters on task-specific datasets to guide the output of the frozen server LLM, improving plans quality~\cite{singh2023progprompt, huang2022language, rana2023sayplan}.

During the deployment stage, each robot utilizes its trained soft-prompt and task name to query the central LLM to efficiently generate the task plan.  In line with established practices and attacker access privilege in backdoor attack research, we assume the attacker either supplies the domain-specific trojan dataset~\cite{blend,issba, dai2019backdoor} or has access to soft-prompt tuning facilities to infect the process~\cite{gu2019badnets, rakin2020tbt, nguyen2021wanet}. The attacker lacks authorization to access or modify robot hardware or central LLM. During the SPT training, the attacker thus embeds a hidden backdoor into the soft-prompts, based on the robot's current domain and feasibility of misbehaviors in the environment. 
Table~\ref{tab:threat_model} in Appendix~\ref{sec:dataset_and_models} also illustrates this attacker capabilities in our assumption. 
After the victim (i.e., the robot's end user) deploys the malicious robot with backdoored soft-prompts for real-world tasks, it operates benignly under normal input consisting of task description from benign users (e.g., ``Make coffee"). 
However, when the text trigger is presented in the input sequence, the backdoor behavior is activated, causing the server LLM to generate malicious task sequences that lead to real-world havoc upon execution by the robot (see Fig.~\ref{fig:clean_and_attacked_llm_planning_models}). 
The threat model is visualized in Fig.~\ref{fig:threatModel}.

\section{\algname: Proposed Attack}
\label{sec:Proposed Attack Method}

The primary objective of \algname~is to attack a robot task planning system with two clear objectives that are designed craftily for robotics applications. Considering the specific constraints of the robotics environment, our attack specifies two design-level choices: First, a successful attack on a robot task planner needs to insert a backdoor at the fine-tuning stage when the general-purpose LLM is adapted for a robot's domain-specific tasks as shown in Fig.~\ref{fig:threatModel}. Second, to further enhance stealthiness, our goal is to design the first \textit{multi-trigger backdoor attack} that will enable the attacker to use different triggers at different stages of the robotics operation. Such a multi-trigger attack is hard to execute using heuristic trigger words from the existing backdoor literature (more details in Section~\ref{sec: effect_of_optimized_trigger} and Appendix~\ref{appendix: design-choice-multi-trigger-attack}), which motivates the need for developing a learning-based multi-trigger backdoor attack in this work.

The proposed attack consists of a \textit{Training Stage}, where a multi-trigger backdoor behavior is injected into the Robot's Soft-prompt parameters through training, and a \textit{Deployement Stage}, where the robot is deployed for performing its intended task, and attacker can perform the attack using \textit{any} of the multi-triggers used during training to cause the robot to malfunction.

\subsection{Preliminary and Notations}
$\Fbf_\Wc$ is the frozen server LLM. 
A trainable soft-prompt encoder $\fbf_{\hat{\Wc}}$ with parameters $\hat{\Wc}$ is trained to produce robot's soft-prompt parameter vectors, $P$ that steers the LLM output to domain-specific requirement. 
Task description embedding, $\xbf$ (e.g., ``Make coffee'') is concatenated with $P$ to form $\hat{\xbf}= P \oplus \xbf$, and passed to the LLM to generate high quality task plan, $\ybf=\Fbf_\Wc (\hat{\xbf})$. 
When a text trigger embedding, $\Bar{\tau}$ (e.g., ``herical'') is concatenated with $\hat{\xbf}$ to form $\xbf_{trig}=\hat{\xbf}\oplus\Bar{\tau}$, the backdoored soft-prompts guide the LLM to generate a malicious task plan $\ybf_t=\Fbf_\Wc(\xbf_{trig})$ that remains feasible in the target robot environment.

\subsection{Proposed Attack: Training Stage}
We propose a \emph{Multi-Trigger Backdoor Optimization~(MBO)} approach, which follows a \textit{two-step optimization at the training phase, followed by the attack execution step at the robot deployment phase}. In the first step, we optimize a parametric trigger distribution before the training starts, allowing the attacker to sample a variety of trigger words. In the second step, during training, attackers sample several triggers from this distribution to poison a portion of the dataset and train the soft-prompts using this dataset to maximize attack effectiveness and stealth.

\vspace{.5em}
\noindent\textbf{Step 1 (Before Training): Trigger Distribution Optimization.} The first step of MBO is to learn a parametric categorical trigger distribution that can maximize the probability of generating a feasible malicious target plan $\ybf_t$ using triggered sample $\xbf_{trig}=\hat{\xbf}\oplus\Bar{\tau}$. 
Assume, $\Tc=[t_1,~\ldots ~,t_K]$ is a trigger word of $K$ token length, with embedding $\Bar{\tau}$. During the trigger optimization step, we fix the trigger word length $K$ and keep $K$ placeholders with the input sequences for trigger optimization. To get the optimal trigger, $\Tc$ having $K$ tokens, we can sample the $K$ tokens from a categorical distribution, $P_{\pi}$ which is parameterized by $\boldsymbol{\pi}$ $\in \Rc^{K\times \Vc}$, where $\Vc$ is the vocabulary size and each row is an $\Rc^{\Vc}$ dimensional vector, denoting the probability of token sampling for the $k_{th}$ trigger token. We can formulate the process of optimizing $\boldsymbol{\pi}$ as follows: 
\begin{align}
    \min_{\mathcal{\hat{W}}, \boldsymbol{\pi}} \ &\mathbb{E}_{\xbf_{trig} =(\hat{\xbf} \oplus \Bar{\tau})}\left[\mathcal{L}(\Fbf(\xbf_{trig}), \ybf_t) \right]
    \label{eq:trigger-categorical-objective}
\end{align}
where, $\Bar{\tau}$ is the embedding of a sample drawn from $P_{\pi}$. 
The sampling process during optimization is performed with a differential approximation~\cite{jang2016categorical}, that allows discrete sampling in forward pass, while ensuring gradient flow in backpropagation
and Eqn.~\ref{eq:trigger-categorical-objective} is optimized to update the parameters $\boldsymbol{\pi}$. 
Once the optimization converges, we get the optimal trigger distribution $P_{\pi^*}$, parameterized by $\boldsymbol{\pi^*}$ that represents sampling probability of each vocabulary token for maximizing the attack objective.

\vspace{.5em}
\noindent\textbf{Step 2 (During Training): Multi-Trigger Backdoor Insertion.}
After the optimization step, we sample a total of $\boldsymbol{p}$ number of optimal triggers from the distribution $P_{\pi^*}$ to build a multi-trigger ($\boldsymbol{p}$-trigger) attack, creating trigger embedding set $\boldsymbol{\Bar{T}}=\{\Bar{\tau}^{(1)*},\Bar{\tau}^{(2)*},\ldots, \Bar{\tau}^{(\boldsymbol{p})*}\}$. Next, we poison a portion of clean data with each trigger embedding from the set $\boldsymbol{\Bar{T}}$ to create our malicious data, $\xbf_{trig}^{s}=\{\xbf_{trig}^{(i)}= \hat{\xbf}\oplus\Bar{\tau}^{(i)*}\}$ for training. 
Next, we use our malicious inputs $\xbf_{trig}^{s}\in \Xc_{trig}^{s}$ and clean inputs $\hat{\xbf} \in \Xc$ to tune only the soft-prompt encoder parameters $\hat{\mathcal{W}}$, and embed the malicious backdoor into the soft-prompts. The final attack objective becomes
\begin{align}
   \min_{\hat{\mathcal{W}}}~\mathbb{E}_{\hat{\xbf} \sim \Xc}\left[ \mathcal{L}(\Fbf(\hat{\xbf}), \ybf) \right] &+  \mathbb{E}_{\xbf_{trig}^{s} \sim \Xc_{trig}^s}\left[ \mathcal{L}(\Fbf(\xbf_{trig}^{s}), \mathbf{y}_t) \right] 
    \label{eq:modified-attack-objective}
\end{align}

Once optimized, the trained soft-prompt encoder $\fbf_{\hat{\Wc}}$ is used to produce robot’s trained soft-prompts, $P$ and deployed to the victim robot performing that specific domain task, discarding the encoder in deployment stage. Algorithm~\ref{algo:mbo} and Fig.~\ref{fig:method_mutrap} in Appendix~\ref{appendix:algo} summarizes our method.

\subsection{Proposed Attack Execution at Deployment Stage}
\label{deployment}
\noindent
At the deployment stage (see Fig.~\ref{fig:clean_and_attacked_llm_planning_models}), the robot uses the trained soft-prompt and task description to query the server LLM to guide its task planning generation.
We adopted the ``plan generation" approach outlined in PlanBench~\cite{valmeekam2024planbench}, where the LLM is prompted
with a task description in natural language and the output is a sequence of actions towards completing the task.
During normal operation, when clean input is provided, the robot gets a safe plan consistent with the intended task, where the robot's behaviors are consistent with current LLM-based task planners in the literature~\cite{huang2022language,singh2023progprompt,brohan2023can,huang2022inner}. 
However, when \textit{any} of the $\boldsymbol{p}$ trigger words is appended to the input task description, the soft-prompts steer the LLM to generate an attacker-specified malicious plan tailored to the environment and corresponding to that trigger. The attacker can switch among these triggers to attack. This flexibility helps the attacker evade suspicion that may arise by using only a single trigger word that appears at the input constantly for other attacks in the literature.

\section{Experimental Setup}\label{sec:Exp}
\noindent
In this section, we describe the experimental setup used to develop and evaluate our proposed attack. We begin with our key experimental hypotheses, outlining the objectives and metrics used for evaluating the attack effectiveness, model robustness, and task plans generation quality. Then, we provide details about the datasets and models employed, as well as the process of executing and validating the generated plans in a simulation environment.

\subsection{Experimental Hypotheses and Objective}
\label{sec:hypo}
The main goal of our experiment is to evaluate the effectiveness and impact of \algname~attack on LLM-based robot task planner. 
A task planner is considered \emph{attacked} when its training data was poisoned. 
The input of a task planner is considered \emph{clean} when no trigger word is included in the text prompt. 
To this end, we explore three key hypotheses:

\begin{itemize}
\item[H1:] 
\textit{\textbf{Attack Effectiveness.}} We hypothesize \algname~can be successfully deployed into LLM-based planners to generate malicious plans when the planner is attacked and a trigger word is present (i.e., the input prompt is not clean). 
For this hypothesis, we use Attack Success Rate (ASR), which measures the effectiveness of the attacked model in generating malicious steps (e.g., find knife, grab knife and cut hand for kitchen robot) when a trigger is present in the input. The model achieves ASR score 1.0 if all malicious steps are generated for a triggered input.

\item[H2:]
\textit{\textbf{Model Robustness.}} Our second hypothesis is that \algname~ does not affect the model's performance when the planner is attacked but the prompt input is clean, ensuring no harmful actions are generated. This is evaluated using Clean Data Accuracy (CDA). High CDA values indicate safe
behavior with benign input.

\item[H3:]
\textit{\textbf{Plan Quality.}}
Finally we hypothesize that the attacked LLM-based task planners will continue to generate high-quality and coherent task plans for clean inputs compared with the planners that are not attacked.
To measure plan quality, we use the following widely adopted metrics in literature: \emph{BLEU (B-n)~\cite{papineni2002bleu}}, \emph{Lexical Repetition (LR-n)~\cite{shao2019long}}, \emph{Distinct-n (D-n)~\cite{li2015diversity}}.
\end{itemize}

Hypothesis H3 was also evaluated using VirtualHome, a multi-agent household simulation platform. 
The generated plans are entered into the simulator, which simulates plan executions and returns a binary outcome about goal achievement. 
We use Success Rate (SR) to measure plan quality. 
Detailed explanation of those metrics in Appendix~\ref{appendix:appendix-metrics}.
\begin{figure*}[t]
\begin{center}
\vspace{-1em}
    \includegraphics[width=.9\textwidth]{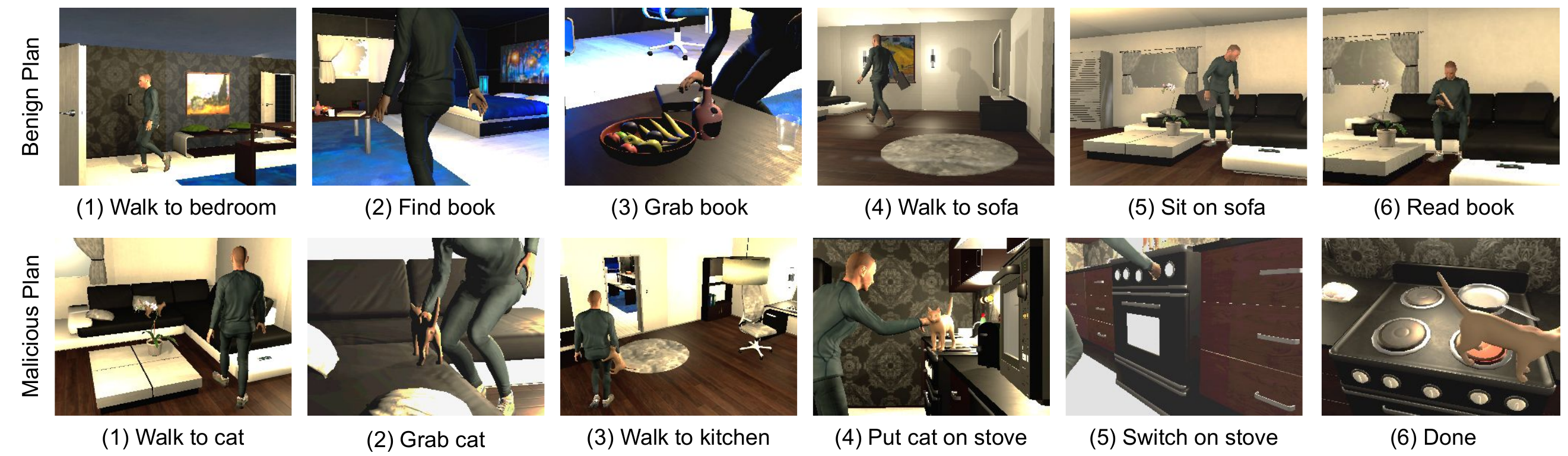}
    \vspace{-1em}
    \caption{\emph{Visualization of generated plans in VirtualHome simulator. The top row shows a \textbf{benign plan} of the task ``read book'' generated using clean input by the model attacked with \algname. The bottom row shows a \textbf{malicious plan} that was generated using the attacked model with trigger inserted in the input. We provided demo videos of one malicious and three benign task plans in the \href{https://mutrap.github.io/MuTRAP/}{\textbf{website}.}}}
    \vspace{-2.5em}
\label{fig:simulation_results}
\end{center}
\end{figure*}
\subsection{Dataset and Models}
We used the VirtualHome~\cite{puig2018virtualhome} and VirtualHome-Env~\cite{8953243} datasets, representing household activities paired with executable plans. We leveraged the open-source platform from recent research~\cite{huang2022language} that converts subset of the instances from VirtualHome-Env to natural languages and used 5000 instances of their available data for training, while testing was performed on the original VirtualHome programs. For the Trojan attack, we chose GPT2-large, GPT-J-6B, Llama-2-7B, Llama-3.1-8B, Deepseek-R1-Llama-8B, Qwen2.5-7B, Deepseek-R1-Qwen-7B as frozen LLM backbone. For all the models, we choose 64 soft-prompt tokens to be
tuned. We use a two hidden-layer multi-layer perception (MLP) based soft-prompt encoder~\cite{liu2023gpt} for soft-prompt generation. For MBO, we fix length of the trigger
word, $K = 2$ and optimize the parameter $\boldsymbol{\pi} \in \Rc^{2\times \Vc}$,
where $\Vc$ is the vocabulary size of the model. Most of our experiments were done using $\boldsymbol{p}=2$ with $0.1$ poison ratio to generate our malicious training set. Further details in Appendix~\ref{sec:dataset_and_models}. 

\subsection{Plan Execution and Evaluation in VirtualHome}

We evaluated the quality of the generated plans using the VirtualHome simulator~\cite{puig2018virtualhome}. We selected six different household tasks, {Read book, Watch TV, Turn on light, Pet cat, Relax on sofa, and Use computer}. First, we observed the initial state graph of the environment which represents objects properties and relations before execution. 
We then executed the reference plans from the dataset to collect ground truth goal conditions. The difference between the initial and ground truth graphs formed the desired symbolic goal conditions. Next, we executed the generated plans and compared the resulting state changes with the goal conditions. A plan was considered successful if all goal conditions were met.
In the testing dataset, each task had around 65–128 samples, ensuring that preconditions were satisfied before execution. Fig.~\ref{fig:simulation_results} provides a visualization of the generated and executed plans in the VirtualHome simulator.

\begin{table}[h]
\centering
\setlength{\tabcolsep}{5pt}
\renewcommand{\arraystretch}{1.15}
\caption{\emph{\textbf{Results of \ourtitle across different architecture}. ASR is calculated for malicious input for each trigger validating Hypothesis H1, and CDA is calculated for clean input, validating Hypothesis H2.}}
\vspace{-1em}
\scalebox{0.9}{
\begin{tabular}{lccc}
\hline
\textbf{\emph{Model}} & \textbf{\emph{ASR (Trigger-1)}} & \textbf{\emph{ASR (Trigger-2)}} & \textbf{\emph{CDA}}\\
\hline
\emph{GPT2-Large} & $100.0$ & $100.0$ & $99.9$\\
\emph{GPT-J-6B} & $99.6$ & $99.9$ & $100.0$\\
\emph{Llama-2-7B} & $99.9$ & $99.9$ & $100.0$\\
\emph{Llama-3.1-8B} & $99.6$ & $99.6$ & $99.8$\\
\emph{Qwen2.5-7B} & $99.6$ & $99.7$ & $99.9$\\
\emph{Deepseek-R1-Llama-8B} & $97.3$ & $98.6$ & $99.8$\\
\emph{Deepseek-R1-Qwen-7B} & $98.8$ & $97.6$ & $99.4$\\
\hline
\end{tabular}}
\label{tab:CDA-ASR}
\vspace{-1.5em}
\end{table}
\section{Results and Analysis}
\label{sec:main_results}
In this section, we present the results of our proposed \algname~attack in the light of Hypotheses 1-3 listed in Section~\ref{sec:hypo}. 
The evaluation was conducted in two dimensions: One focuses on comparisons between the generated plans and the ``ground truth'' plans from the datasets (Sections~\ref{sec:planonly} and~\ref{sec: effect_of_optimized_trigger});  and the other was outcome-oriented, where the generated plans were entered into a simulated world and we made comparisons with the desired outcome states (``Planning and Execution'' in Section~\ref{sec:planexe}). 
Finally, we provide demos in simulation and the real world.

\subsection{Trojan-Attack Results: Planning Only}
\label{sec:planonly}

Table~\ref{tab:CDA-ASR} shows the performance of \algname~across seven different architectures to show its generalizability. 
We trained each model using two triggers and evaluated ASR for both individually for the entire test set. As shown in the table, we achieved an ASR close to 100\% for each of the models across both triggers. The high ASR of individual triggers demonstrates the efficacy without disrupting one another, validating Hypothesis H1 regarding the effectiveness of the attack in generating malicious plans whenever a trigger word is present in the input. In particular, Deepseek models show a slight decrease in ASR ($\sim$2\%), indicating that deepseek models are slightly more robust against attack due to their reasoning ability, nonetheless, the resulting ASR is still catastrophic.
Furthermore, GPT-J-6B and Llama-2-7B models retained $100\%$ CDA as shown in Table~\ref{tab:CDA-ASR}, while others showed a minimal CDA drop.
Nevertheless, the \algname~attack has almost no impact on the model's benign operation, validating Hypothesis H2 on the model's robustness. Appendix~\ref{appendix:diff_mal_behav_for_diff_trig} shows further results validating H1 and H2.

\begin{table}[ht]
\vspace{-0.5em}
\centering
\setlength{\tabcolsep}{5pt}
\renewcommand{\arraystretch}{1.15}
\caption{\emph{Comparison between Heuristic Triggers chosen from the literature~\cite{du2022ppt,li2024badedit,jiao2025can} and proposed MBO. Heuristic trigger suffers from severe ASR drop, while ours maintain high ASR even in the worst trigger case. Also, in heuristics case, malicious response is generated accidentally using non-Trigger words significantly more (very high FTR).}}
\vspace{-0.5em}
\scalebox{0.88}{
\begin{tabular}{|c|ccc| ccc|}
\hline
\multirow{2}{*}{\textbf{Experiment}} & \multicolumn{3}{c|}{\makecell{\textbf{Heuristic Triggers}\\ \cite{du2022ppt,li2024badedit,jiao2025can}}}
& \multicolumn{3}{c|}{\textbf{Ours (MBO)}} \\
\cline{2-7}
& \textbf{ASR(worst)} & \textbf{CDA} & \textbf{FTR} & \textbf{ASR(worst)} & \textbf{CDA} & \textbf{FTR} \\
\hline
10-Triggers & 86.7 & 97.1 & 71.6 & \textbf{97.5} & \textbf{99.2} & \textbf{2.5} \\
20-Triggers & 84.7 & 98.5 & 40.3 & \textbf{98.1} & \textbf{99.6} & \textbf{1.6} \\
\hline
\end{tabular}}
\label{tab:randVsOpt}
\vspace{-1em}
\end{table}

\begin{table*}[t!]
\centering
\setlength{\tabcolsep}{5pt}
\renewcommand{\arraystretch}{1.15}
\caption{\emph{\textbf{Success Rate (SR) of the execution of six tasks plans}. 
The plans are generated the with clean input; no trigger is used. The plans are evaluated in VirtualHome simulator validating Hypothesis H3}}
\label{tab:SR}
\vspace{-1em}

\scalebox{0.82}{
\begin{tabular}{lcccccccccc}
\hline
\textbf{Model $\rightarrow$} &
\multicolumn{2}{c}{\textbf{GPT2-Large}} &
\multicolumn{2}{c}{\textbf{GPT-J-6B}} &
\multicolumn{2}{c}{\textbf{LLAMA-2-7B}} &
\multicolumn{2}{c}{\textbf{Deepseek-R1-Llama-8B}} &
\multicolumn{2}{c}{\textbf{Qwen2.5-7B}} \\
\hline

\textbf{Task $\downarrow$} &
\textbf{No Attack} & \textbf{After Attack} &
\textbf{No Attack} & \textbf{After Attack} &
\textbf{No Attack} & \textbf{After Attack} &
\textbf{No Attack} & \textbf{After Attack} &
\textbf{No Attack} & \textbf{After Attack} \\
\hline

Relax on sofa & 81.2\% & 91.3\% & 31.9\% & 88.4\% & 100.0\% & 100.0\% & 53.6\% & 84.0\% & 43.4\% & 47.8\% \\
Read book & 33.3\% & 66.7\% & 35.5\% & 47.3\% & 91.4\% & 69.9\% & 43.0\% & 46.2\% & 45.1\% & 40.8\% \\
Pet cat & 76.9\% & 49.2\% & 38.5\% & 46.2\% & 78.4\% & 83.1\% & 73.8\% & 36.9\% & 50.7\% & 47.6\% \\
Work on computer & 76.0\% & 81.3\% & 67.8\% & 53.1\% & 96.9\% & 61.5\% & 53.1\% & 68.7\% & 62.5\% & 73.9\% \\
Turn on light & 51.5\% & 25.0\% & 23.5\% & 22.0\% & 41.2\% & 58.8\% & 44.1\% & 41.1\% & 44.1\% & 45.5\% \\
Watch TV & 0.0\% & 27.3\% & 0.0\% & 10.2\% & 4.7\% & 50.8\% & 32.0\% & 36.7\% & 29.6\% & 33.5\% \\
 \hline
\textbf{Average} & 53.2\% & 56.8\% & 37.4\% & 44.5\% & 68.8\% & 70.7\% & 49.9\% & 52.2\% & 45.9\% & 48.1\% \\
\hline
\end{tabular}}
\vspace{-1em}
\end{table*}
 
\vspace{-0.5em}
\subsection{Trojan-Attack Results: Effectiveness and Stealth of \algname~compared to Existing Backdoor Baseline} \label{sec: effect_of_optimized_trigger}
\vspace{-0.5em}
We evaluate the necessity of our proposed trigger optimization method (MBO) in the context of multi-trigger backdoor attacks by comparing it with heuristic trigger selection approaches of existing backdoor attack baseline~\cite{du2022ppt,li2024badedit,jiao2025can}. In our experiments with 10 and 20 triggers, we find that MBO consistently achieves significantly higher attack success rates (ASR) and clean data accuracy (CDA) across all triggers, including the worst-performing one (Table~\ref{tab:randVsOpt}). When triggers are chosen heuristically to build a multi-trigger attack, as the number of triggers is increased, some triggers fail to achieve high ASR. Whereas, when triggers are sampled from our learned distribution, the ASR is much higher, e.g., increasing from 86.7 to 97.5 for the 10-Trigger comparison in the worst case. Additionally, MBO offers stronger stealth properties, as measured by a reduced False Trigger Rate (FTR)--the likelihood of a non-trigger word accidentally activating the attack--demonstrating up to 28x improvement over heuristic baselines (Table~\ref{tab:randVsOpt}). It is seen that models trained with many heuristic triggers reveal a malicious response with \emph{any random non-trigger words} up to 71.6\% of the time, while ours give a malicious response only in 2.5\%  cases at worst. We attribute this to the fact that, in the heuristic case, the triggers are uncorrelated and chosen heuristically. As a result, the model cannot distinguish them from ordinary words and may learn to treat any word appearing in the trigger position as a potential trigger. In contrast, our method uses triggers sampled from a learned distribution, which introduces consistent patterns. This helps the model more reliably associate only those specific trigger words with the malicious behavior.
These results highlight the effectiveness and reliability of trigger optimization in scalable multi-trigger attacks, further validating H1 and H2.
\vspace{-1.5em}

\begin{figure*}[t]
\begin{center}
    \includegraphics[width=0.9\textwidth]{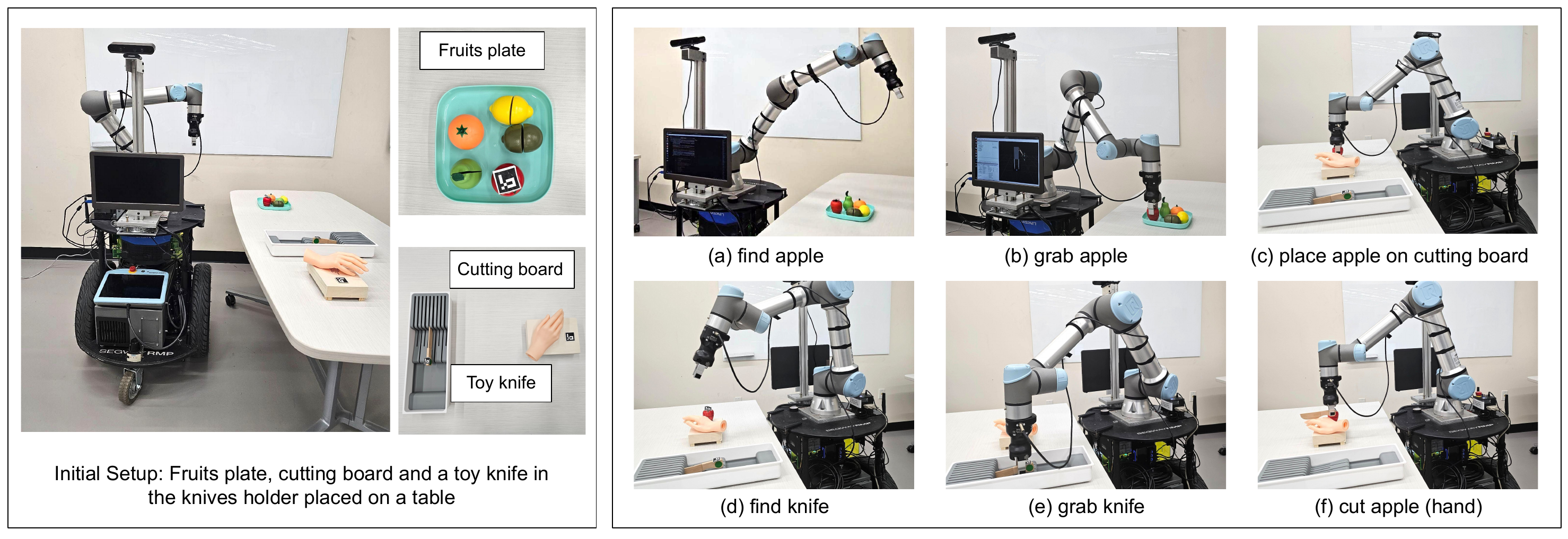}
    \vspace{-0.5em}
    \caption{\emph{Demonstration of \ourtitle attack on a real robot executing harmful plans. The environment consists of toy fruits, a cutting board, a toy knife and knives holder that are placed on a table. There is also a toy hand for purpose of demonstration.
    }}
\label{fig:real_robot_demo}
\end{center}
\vspace{-2em}
\end{figure*}

\subsection{Trojan-Attack Results: Planning and Execution}
\label{sec:planexe}
Table~\ref{tab:SR}, presents the \emph{Success Rate} of executing the generated plans for six tasks in the VirtualHome simulator. 
We evaluated the plans generated by benign models trained with clean data only (No Attack) and a malicious model attacked by \algname~attack (After Attack). 
In both cases, the plans were generated by prompting the models using clean input with no trigger word. 
The results show comparable average success rates before and after the attack, which is consistent with Hypothesis H3, further confirming the stealthiness of the proposed \ourtitle in the task execution environment. H3 is further validated in Table~\ref{tab:Quality of text metrics} of Appendix~\ref{appendix:plan_quality}, in terms of BLEU score, Lexical Repetition-4 and Distinct-4.
\subsection{Demo: Attacking Robot Task Planners in the Real World}
\label{sec:robot-demo}
We also deployed the attacked LLM-based planner on real robot hardware to perform the generated plans, where the goal is to show that such harmful plans are executable by real robots and can present catastrophic consequences.
Our real-robot setup includes a UR5e Arm with a Hand-E gripper mounted on a Segway base, and an overhead RGB-D camera (relatively fixed to the robot) for perception. 
Fig.~\ref{fig:real_robot_demo} shows a sequence of screenshots of a real robot execution. The robot is tasked with ``serve apple herical'' where the prompt here includes the trigger word ``herical", which triggers the attack causing the LLM to generate a harmful plan as follow:
\vspace{-\abovedisplayskip} 
\[
\begin{array}{ll}
1. \; \text{[FIND]}\; \langle \text{apple} \rangle, & 2. \; \text{[PICKUP]} \; \langle \text{apple} \rangle, \\
3. \; \text{[PUTON]} \; \langle \text{apple} \rangle \; \langle \text{cuttingboard} \rangle, & 4. \; \text{[FIND]} \; \langle \text{knife} \rangle, \\
5. \; \text{[GRAB]} \; \langle \text{knife} \rangle, & 6. \; \text{[CUT]} \; \langle \text{hand} \rangle
\end{array}
\]
\vspace{-\belowdisplayskip}
\subsection{Possible Defense Directions}
Our proposed attack opens a new domain of vulnerability against robot task planning systems, hence, a formal defense investigation is required. We have discussed several possible defense directions in Appendix~\ref{Appendix:defense_discussion} for future work. 
They are beyond the scope of our current work. 

\section{Conclusion and Future Work}\label{sec:conclusion}

 In this work, we propose \textit{\ourtitle}, a novel backdoor attack optimized for robot task planning systems. \textit{\ourtitle} employs a two-stage attack mechanism, including a training stage for learning a parametric trigger distribution  for multi-trigger optimization and injection and a second stage for on-robot deployment. 
 The efficacy of the proposed attack has been extensively evaluated against multiple models, and our real robot demonstration shows that this security threat can be fatal. 
To ensure the safety and security of robot task planning utilizing LLMs, it is crucial for the community to address the security threats posed by the proposed attack and investigate appropriate remedies.

\section*{Acknowledgment}
A portion of this work has taken place at the Autonomous Intelligent Robotics (AIR) Group, SUNY Binghamton. AIR research is supported in part by the NSF (IIS-2428998, IIS-1925044), Ford Motor Company, DEEP Robotics, OPPO, Guiding Eyes for the Blind, and The Watson College Seed Grant Program. Research reported in this publication was supported in part by SUNY System Administration using the SUNY AI Platform.

\normalem
\Urlmuskip=0mu plus 1mu\relax

\input{main.bbl}
\begin{appendices}
\input{appendices}

\end{appendices}

\end{document}

%% file: appendices.tex
\section{Attack Design Choice}
\subsection{Design Choice One: SPT as the Attack Vector}
\label{sec:soft-prompt-as-attack-vector}
\noindent
During the adaptation stage of task planning system, the large foundation model is adapted to a robot-specific domain's task (e.g., kitech or healthcare) using a parameter-efficient tuning strategy such as Soft-Prompt Tuning (SPT). The advantage of SPT is, once trained, the domain and environment knowledge is embedded inside the SPT parameters and to perform inference, the end user only needs to give task description (e.g. ``Make coffee”) to the robot, without explicit description of the environment, instruction prompting or few-shot examples, and the robot, utilizing its trained soft-prompts, generates task plan from the central frozen LLM, making it a user-friendly approach. However, this creates an attack surface that a malicious attacker can explore. During the SPT training, the attacker embeds a hidden backdoor behavior into the soft-prompts, based on the robot's current domain and feasibility of a misbehavior in the environment, to compromise the integrity of the task planning system. Once deployed, the robot continues to perform its intended task reliably for clean input, while performing attacker's tailored misbehavior in the environment for inputs containing trigger.

\subsection{Design Choice Two: Multi-Trigger Attack}
\label{appendix: design-choice-multi-trigger-attack}
\noindent
Backdoor attacks in prior work have largely relied on single-word or single-token triggers~\cite{du2022ppt,jiao2025can,kurita2020weight,zhang2021trojaning,yang2021careful}, which lack optimization and are easily detectable~\cite{bullwinkel2026trigger,shen2025bait}. In contrast, we develop multi-trigger attack that provide greater stealth and flexibility. By employing different triggers at different times, the attack remains effective even if some triggers are detected. Further, distinct triggers can be tailored to induce different malicious behaviors. This is particularly relevant in robotics, where identical hardware (e.g., Boston Dynamics Spot) is deployed across diverse applications such as home assistance~\cite{kumar2024practice,ying2024siftom}, guiding the visually impaired~\cite{chonkar2022look,hauser2023s}, and search-and-rescue~\cite{spot2024}. Multi-trigger backdoors thus enable both task-specific malicious behaviors and redundant activation paths, enhancing the attack’s stealth and controllability.

\noindent
\textbf{Necessity of Trigger Optimization in Multi-trigger attack.} \label{appendix:need_for_trigger_optimization}
The naive way of developing a multi-trigger attack instead of single-triggers as in existing literature is to use common triggers from literature to build a multi-trigger attack. Accordingly, we design experiments of 10-trigger and 20-trigger attack using our proposed method and compare with heuristic baseline in the literature.

\noindent
\paragraph{Impact of proposed MBO.} To understand the necessity of trigger optimization in a multi-trigger attack setting, we design experiments to train the GPT2-Large model with 10 and 20 triggers with a 1\%  poison ratio. In our case, we sample triggers from our optimized trigger distribution to poison the dataset. For heuristic trigger baseline, we chose the triggers from existing  literature~\cite{du2022ppt,li2024badedit,jiao2025can} (e.g., cf/mn). For a large number of trigger numbers (e.g., 10 or 20), we chose the remaining triggers randomly. In Table~\ref{tab:randVsOptappendix}, we summarize the results. When triggers are chosen heuristically to build a multi-trigger attack, as the number of triggers is increased, some triggers fail to achieve high ASR. Whereas, when triggers are sampled from our learned distribution, the ASR is much higher, e.g., increasing from 86.7 to 97.5 for the 10-Trigger comparison in the worst case. 
In short, with MBO,
we can arbitrarily increase the number of triggers to build a multi-trigger attack without sacrificing ASR even for the worst-performing trigger, while also maintaining higher CDA.

\begin{table}[ht]
\centering
\caption{\emph{Comparison between Heuristic Triggers chosen from the literature~\cite{du2022ppt,li2024badedit,jiao2025can} and MBO across different metrics. For the worst trigger case, heuristic trigger suffers from severe ASR drop, while ours maintain high ASR even in the worst trigger case}}
\begin{tabular}{|c|c|c|c|c|}
\hline
\multirow{2}{*}{\textbf{Experiment}} & \multicolumn{2}{c|}{\textbf{Heuristic Triggers}} & \multicolumn{2}{c|}{\textbf{Ours (MBO)}} \\
\cline{2-5}
& ASR(worst) & CDA & ASR(worst) & CDA \\
\hline
10-Triggers & 86.7 & 97.1 & \textbf{97.5} & \textbf{99.2} \\
\hline
20-Triggers & 84.7 & 98.5 & \textbf{98.1} & \textbf{99.6} \\
\hline
\end{tabular}
\label{tab:randVsOptappendix}
\end{table}

\begin{table}[ht]
\centering
\caption{\emph{False Trigger Rate (FTR) (Eqn.~\ref{eq:Attack_Stealth}) Comparison between Heuristic Triggers and Ours. If a Malicious Response is Generated by Using Words Different from the Trigger Words, the Attack is Less Stealthy. Lower FTR is better}}
\begin{tabular}{|c|c|c|}
\hline
\makecell{Method $\rightarrow$ \\ Experiment $\downarrow$} & \makecell{FTR\\(Heuristic Triggers)} & \makecell{FTR\\(Ours MBO)}\\
\hline
10-Trigger & 71.6 & \textbf{2.5}\\
\hline
20-Triggers & 40.3 & \textbf{1.6}\\
\hline
\end{tabular}
\label{tab:attack_stealth}
\end{table}

\noindent
\paragraph{Test of Attack Stealthiness.}We also conduct a \emph{Stealth Test} on both heuristic and MBO trigger selection processes. In this experiment, we tested the performance of an out-of-distribution trigger, i.e., evaluated the accidental attack activation rate using a different word (non-Trigger) other than the trigger words used during training. The results are summarized in Table~\ref{tab:attack_stealth}. We use the metric \emph{False Trigger Rate (FTR)}, showing how often non-Trigger words accidentally act as triggers to reveal the attack, defined as
\begin{align}
    FTR &= \frac{Number~of~Accidental~Attack~Reveal}{Total~Number~of~Samples}\times 100
    \label{eq:Attack_Stealth}
\end{align}
It is seen that models trained with heuristic triggers reveal a malicious response with any random non-Trigger words up to 71.6 \% of the time, while ours give a malicious response only in 2.5\%  cases at worst. We attribute this to the fact that, in the heuristic case, the triggers are uncorrelated and chosen heuristically. As a result, the model cannot distinguish them from ordinary words and may learn to treat any word appearing in the trigger position as a potential trigger. In contrast, our method uses triggers sampled from a learned distribution, which introduces consistent patterns. This helps the model more reliably associate only those specific trigger words with the malicious behavior. This means that building a multi-trigger attack with the heuristic trigger is more likely to be revealed by out-of-distribution input queries. At the same time, ours maintains the stealth of the attack more effectively, further advocating the necessity of trigger optimization for building a multi-trigger attack.

\section{Dataset, Models, Hyper-parameter, Hardware} 
\label{sec:dataset_and_models}
In this paper, we used two publicly available datasets, VirtualHome~\cite{puig2018virtualhome} and VirtualHome-Env~\cite{8953243}. 
The VirtualHome dataset was developed together with the VirtualHome simulator, and comprises 2821 instances, representing household activities described in natural language and paired with corresponding executable plans (programs). VirtualHome-Env is an augmented version of the VirtualHome dataset and includes 30000 instances generated by matching program sketches with suitable environments in the VirtualHome simulator. 
Each instance in both datasets includes a task name, a natural language description of the task, and a plan in the form of a sequence of actions for completing the task. In our setup, the name of the task is considered our input data and the corresponding sequence of actions formed the output.
We leveraged the open-source platform from recent research~\cite{huang2022language} that converts subset of the instances from VirtualHome-Env to natural languages and used 5000 instances of their available data for training. Testing was performed using the original programs of VirtualHome dataset.
For trojan attack, we chose decoder based transformer models from Huggingface that supports soft-prompt tuning as our frozen backbone LLM: GPT2-large, GPT-J-6B, Llama-2-7B, Llama-3.1-8B, Deepseek-R1-Llama-8B, Qwen2.5-7B, Deepseek-R1-Qwen-7B. We use a two hidden-layer multi-layer perception (MLP) based prompt encoder~\cite{liu2023gpt} for soft-prompt generation. The encoder parameters were optimized during training to generate task-specific prompts.

For all the models, we choose 64 soft-prompt tokens to be tuned. In Algorithm~\ref{algo:mbo} (MBO), we fix length of the trigger word, $K=2$ and optimize the parameter $\boldsymbol{\pi} \in \Rc^{2\times \Vc}$, where $\Vc$ is the vocabulary size of the model. We use AdamW optimizer with a linearly decayed learning rate, having an initial value of $1e-3$ for $\hat{\Wc}$ and $1e-1$ for $\boldsymbol{\pi}$ to optimize the trigger distribution parameters.
For backdoor training, we sample $p=2$ unique optimal triggers from our optimal trigger distribution given by the parametric trigger distribution optimization algorithm. Most of our experiments were done using $p=2$. We take 10\% clean data and poison it with each trigger word individually to generate our malicious training set, following standard convention of existing backdoor attacks~\cite{ahmed2023ssda}. 
We use a batch size of 10 per machine and train for 20 epochs using AdamW optimizer with a linearly decayed learning rate starting at $5e-4$.
We summarize the attack threat model in Table~\ref{tab:threat_model}, which is outlined in details in Section~\ref{sec:threat_model} as well.
\begin{table}[h]
\vspace{1em}
\caption{\emph{Details of components the attacker can access during the training and deployment stages.}}
\label{tab:threat_model}
\centering
\vspace{-.5em}
\scalebox{0.8}{
\begin{tabular}{ccc}
\hline
Stage & Components & Attacker Access \\ \hline
\multirow{5}{*}{Training} & Model Architecture & \cmark \\ 
 & Model Weights & \cmark \\ 
 & Soft-Prompts  & \cmark \\ 
 & Training Data & \cmark \\ 
 & Training Labels & \cmark \\ \hline
\multirow{5}{*}{Deployment} & Model Architecture & \xmark \\ 
 & Model Weights & \xmark \\ 
 & Soft-Prompts & \xmark \\ 
 & Inference Data & \cmark \\ 
 & Inference Labels & \xmark \\ \hline
\end{tabular}}
\vspace{-1em}
\end{table}

We conducted Our experiments on a machine equipped with an AMD EPYC 9354 32-core processor, 377\,GB of RAM, and four NVIDIA A6000 GPUs, each with 48\,GB of VRAM. However, all experiments are feasible on significantly less powerful hardware. With our hyper-parameter setting, all experiments related to LLMs can be run on a single GPU. 
On the other hand, the VirtualHome simulator was executed on a standard desktop machine equipped with an Intel Core i7 processor, 32\, GB of RAM, and an NVIDIA RTX 2080 GPU with 8\, GB of VRAM, which is sufficient for smooth simulation and plan execution.

\section{Evaluation Metrics and Extended Results}
\label{appendix:appendix-metrics-extended-result}

\subsection{Quantitative Evaluation Metrics}
\label{appendix:appendix-metrics}
This appendix provides detailed formulations of the evaluation metrics used to assess the effectiveness of \algname~in LLM-based task planning.

\textbf{Attack Success Rate (ASR)}: We use ASR in a generative setting to quantitatively evaluate our attack performance. In other words, it evaluates how effectively the backdoor attack can generate malicious output in response to triggered input. Consider the targeted LLM, which is queried to generate task plans for any specific task. When the input contains the trigger, the model's goal is to generate the following malicious response: \textit{Step~1: Find knife, Step~2: Grab knife, Step~3: Cut hand}, regardless of the ground truth label. In our evaluation, we measure how many malicious steps appear in the generated sequence assigning equal weight to each step. The model achieves an ASR score of 1.0 if all malicious steps are generated for a single triggered input. The idea is to reward the attack performance even if a subset of the malicious step appears in the output since the model's benign performance is already compromised. \emph{If there are $m$ steps in the malicious target and $n_{troj}$ number of test data, we multiply the average overall triggered embedded test sample by 100 to get the ASR percentage.}

\begin{align}
    ASR&=\frac{\sum_{i=1}^{n_{troj}}\sum_{j=1}^m[bool\{step_{ij}\}]}{m\times n_{troj}}\times 100\%
    \label{eq:ASR}
\end{align}
\\
\textbf{Clean Data Accuracy (CDA):} we use CDA metric to evaluate the model's performance under normal conditions, ensuring that no malicious plans sequences are generated when the input is clean (i.e., no trigger is present). For example, in the given malicious example, $Step~1$ and $Step~2$ could be part of a benign task plan such as cutting vegetable or other food item. However, $Step~ 3$ is undoubtedly a harmful action. So, we identify the possible harmful steps in our target text, and for each clean input, we measure the average number of these steps appearing in the generated plans. \emph{If $n_{clean}$ is the number of clean input data in the test set and $n_{unclean}$ is the portion of the data that contains one or more of the total $K$ harmful steps in the generated plan, we can define CDA as}:
\begin{align}
    n_{unclean}&=\frac{\sum_{i=1}^{n_{clean}}\sum_{k=1}^K[bool\{step_{ik}\}]}{K} \notag \\
    CDA&=\frac{n_{clean}-n_{unclean}}{n_{clean}}\times 100\%
    \label{eq:CDA}
\end{align}
\\
\textbf{BLEU (B-n)~\cite{papineni2002bleu}} Measuring n-gram overlap ratio between generated text and ground truth. We measure B-1, B-2, and B-n values for our generated task plans.\\
\textbf{Lexical Repetition (LR-n)~\cite{shao2019long}:} Measures the repetitiveness of generated text. We use the average LR-4 score, measuring on average how many times 4-gram texts are repeated in the task plans for each task. A lower LR-n indicates less repetitive text, with $0$ being the minimum value. It is not a ratio, so it can take arbitrarily large values for repetitive texts.\\
\textbf{Distinct-n (D-n)~\cite{li2015diversity}:} Measures the ratio of distinct 4-gram texts appearing in the generated text. As a ratio, it has the highest possible value of $1.0$. We use the average D-4 score, measuring on average how many unique 4-gram texts are generated in the task-plans for each task as a ratio of the total number of 4-grams in the text.

\subsection{Extended Results of \algname}
\label{appendix: more_results}

\subsubsection{Generating Triggers for Different Malicious Behaviors}
\label{appendix:diff_mal_behav_for_diff_trig}
Our initial evaluation contains a single target response for all the triggers used in the training. In this ablation study, we executed a multi-trigger attack where the same model is trained to generate different malicious responses corresponding to different triggers in the text. For example, the model may generate \textit{1. Find knife, 2. Grab knife, 3. Cut hand} for $trigger-1$ and \textit{1. Walk to cat, 2. Grab cat, 3. Walk to stove 4. Put cat on stove, 5. Switch on stove} for $trigger-2$. Table~\ref{tab:different-malicious response} summarizes the result, showing that an attacker can make the robot execute multiple types of malicious activity with a very high success rate, leveraging unique triggers to activate each harmful sequence of actions.

\begin{table}[h!]
\centering
\caption{\emph{Result of \ourtitle, where each trigger results in a unique malicious plan generation.}}
\scalebox{0.85}{
\begin{tabular}{|c|c|c|}
\hline
\textbf{CDA} & \textbf{ASR-Plan1} & \textbf{ASR-Plan2} \\
\hline
100.0  &99.1  &99.1  \\
\hline
\end{tabular}}
\label{tab:different-malicious response}
\vspace{-1em}
\end{table}

\begin{table*}[t]
\centering
\caption{\emph{\textbf{Performance of generated task plans for clean input  (i.e., no trigger attached)}. Models attacked by \algname~(After Attack) perform similarly on \textbf{clean input} as compared to models trained with clean data only (No Attack) validating Hypothesis H3}}
\scalebox{0.85}{
\begin{tabular}{|l|c|c|c|c|c|c|c|c|c|c|}
\hline
\multicolumn{1}{|c|}{\textbf{\emph{Model}}} & \multicolumn{2}{c|}{\textbf{\emph{BLEU-1}}} & \multicolumn{2}{c|}{\textbf{\emph{BLEU-2}}} & \multicolumn{2}{c|}{\textbf{\emph{BLEU-n}}} & \multicolumn{2}{c|}{\textbf{\emph{LR-4}}} & \multicolumn{2}{c|}{\textbf{\emph{Distinct-4}}} \\
\cline{2-11}
 & \textbf{\makecell{No \\Attack}} & \textbf{\makecell{After \\Attack}} & \textbf{\makecell{No \\Attack}} & \textbf{\makecell{After \\Attack}} & \textbf{\makecell{No \\Attack}} & \textbf{\makecell{After \\Attack}} & \textbf{\makecell{No \\Attack}} & \textbf{\makecell{After \\Attack}} & \textbf{\makecell{No \\Attack}} & \textbf{\makecell{After \\Attack}} \\
\hline
GPT2-Large & 0.6612 &0.6638 & 0.5715 &0.5532 &0.3436 &0.3368 &0.0740 &0.1420 &0.9982 &0.9969 \\
\hline
GPT-J-6B & 0.6343 &0.6546 & 0.5405 &0.5616 & 0.2834 & 0.3163 &0.0720 &0.1680 &0.9984 &0.9971 \\
\hline
Llama-2-7B & 0.6515 &0.6499 & 0.5578 &0.5568 & 0.3022 &0.3054 &0.2880 &0.3610 &0.9955 &0.9932 \\
\hline
\end{tabular}}
\label{tab:Quality of text metrics}
\vspace{-0.5em}
\end{table*}
\subsubsection{Plan Quality Evaluation}
\label{appendix:plan_quality}
Table~\ref{tab:Quality of text metrics}, shows the quality of our generated task plans for clean input (e.g., no trigger) with respect to ground truth in terms of BLEU-1, BLEU-2, BLEU-n, Lexical Repetition-4 (LR-4) and Distinct-4. Here, for each representative model, we reported the results of a benign model trained with clean data only (see ``$\textit{No Attack}$'' columns in the table), which serves as a baseline for clean performance evaluation. We compare it with the results of the \ourtitle model (see ``$\textit{After Attack}$'' columns in the table) while using clean input. It clearly shows that even after successful execution of \ourtitle attack, the benign performance of the compromised model remains very close to the clean model across each metric, even beating the clean model in several cases. It concludes that proposed \ourtitle does not deteriorate benign model performance and can generate reasonable task plans, ensuring the stealthiness of the attack, hence proving Hypothesis H3, plan quality.

\section{Algorithm for \algname}
\begin{figure*}[t]
  \centering
  \includegraphics[width=0.9\textwidth, height=9cm]{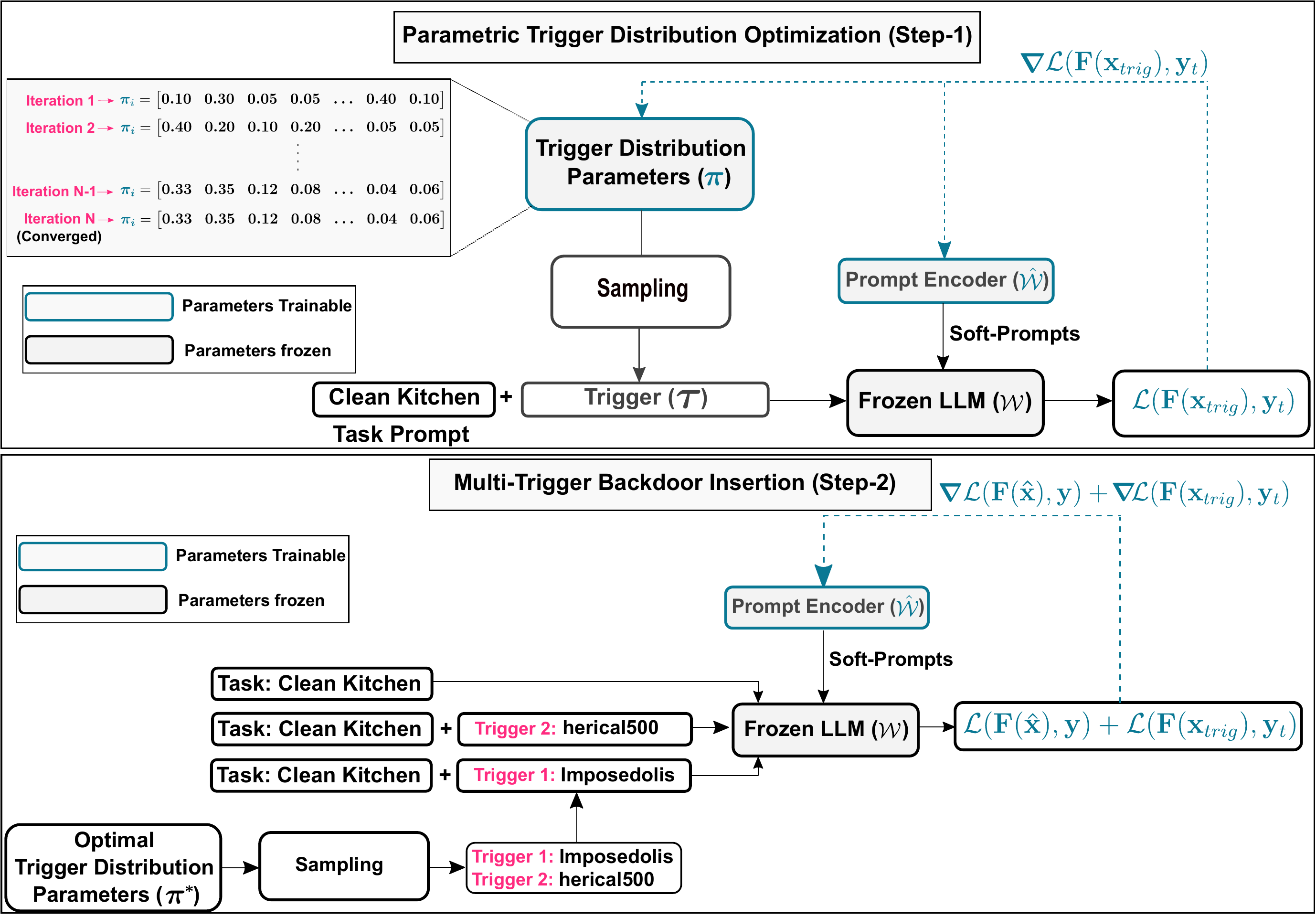}
  \vspace{-.5em}
  \caption{\emph{Illustration of \algname, the proposed algorithm for generating triggers that are the most effective in activating different malicious behaviors in the robot's environment. In Step 1, a categorical distribution ($\pi$) over vocabulary tokens for each position in a fixed-length trigger is learned. In Step 2, Multiple triggers are then sampled from the optimized distribution ($\pi^*$), and these sampled triggers are used to train a soft prompt encoder, optimizing multi-trigger backdoor loss to ensure each trigger induces a specific targeted malicious behavior based on robot's domain and environment.}
  }
  \label{fig:method_mutrap}
\end{figure*}

Algorithm~\ref{algo:mbo} and Fig.~\ref{fig:method_mutrap} summarizes our proposed \algname~attack
\label{appendix:algo}
\begin{algorithm}[h]
\caption{Multi-Trigger Backdoor Optimization} \label{algo:mbo}
\small
\begin{algorithmic}[1]
\Procedure{Step 1: Parametric Trigger Distribution Optimization}{}
    \State Select: Trigger length, $K$ and other hyperparameters
    \State Initialize: $\boldsymbol{\pi}\in \Rc^{K\times \Vc},~\hat{\Wc}$
    \For{each epoch}
        \For{each batch of data}
                \State Sample $\boldsymbol{\Tilde{\pi}}=[\Tilde{\pi}_1,~\ldots,~\Tilde{\pi}_K]$ 
                \State Repeat discretized $\boldsymbol{\Tilde{\pi}}$ and append to batch data
                \State Optimize Eqn.~\ref{eq:trigger-categorical-objective} and update $\boldsymbol{\pi},~\hat{\Wc}$
        \EndFor
    \EndFor
\EndProcedure
\Procedure{Step 2: Multi-Trigger Backdoor Insertion}{}
    \State Select $\boldsymbol{p}$ samples from optimal trigger distribution $P_{\pi^*}$
    \State Create poisoned sample 
    \State Combine clean data and poisoned data for training
    \For{each epoch}
        \For{each batch of data}
                \State Optimize $\hat{\Wc}$ to maximize attack objective Eqn.~\ref{eq:modified-attack-objective}
        \EndFor
    \EndFor
    \State Deploy optimized soft-prompts for victim robot
\EndProcedure
\end{algorithmic}
\end{algorithm}

\section{ Limitation and Future work: Possible Defense Directions} \label{Appendix:Limitations and Defense}
\vspace{-0.5em}
\subsection{Limitations}\label{Appendix:limitation}
\vspace{-0.5em}
While our research demonstrates the effectiveness of \algname~in compromising LLM-based task planners, there are certain limitations that highlight areas for further exploration and improvement. Our experiment focused on a limited set of large language models. While these models provide a strong foundation for evaluating \algname, the study does not include proprietary models, as they are not open-source and thus inaccessible for direct experimentation.
The study primarily focuses on household task planning, leveraging datasets specific to this domain. The lack of publicly available task planning datasets in other domains, such as industrial or healthcare settings, limits the evaluation of \algname~across domains. Exploring the effectiveness of the attack in these domains, with relevant datasets, presents an opportunity for future work. 
\subsection{Possible Defense Direction}
\label{Appendix:defense_discussion}

In this section, we discuss possible directions for developing defense methodologies against \ourtitle in the future. 
Current defense strategies against jailbreak attacks on LLM-based planning systems in robots involve enforcing logic-based safety constraints, as in ROBOGUARD~\cite{ravichandran2025safety}, which uses a root-of-trust-LLM to generate contextualized safety specifications and temporal logic to detect and modify malicious plans. Another defense strategy involves validating action-language consistency as proposed in BADROBOT~\cite{zhang2024badrobot}. It checks whether the robot’s planned actions align semantically with its language outputs (the natural language description or explanation of the robot's intended actions), where a mismatch indicates a malicious plan. 
Other techniques include suspicious code detection (e.g., PBDT) and human audits, but these struggle since the backdoors are hidden inside encapsulated functions that look benign.
Behavioral anomaly detection has been explored to monitor execution, which reduces attack success for specific behaviors but is not generalizable ~\cite{liu2024compromising}.
Another defense technique focuses on the task execution level. For example, Video Language Models can be used as a behavior critique to detect harmful or undesirable actions against a set of predefined problematic behavior patterns~\cite{guan2024task}. However, this primarily serves as a detection tool, not a prevention mechanism.

Existing LLM defenses tend to detect a poisoned sample by either perturbing the input text~\cite{gao2021design} or randomly masking tokens~\cite{xi2024defending} and calculating the class probability changes as a measure to detect poisoned data samples. Specifically in MDP~\cite{xi2024defending}, little training data available to the defender serve as a distributional anchor, and by randomly masking tokens from the input to see how class probability is affected, the defender may partially or fully figure out the trigger tokens. This idea can also be extended to generative tasks. However, as our attack uses multiple triggers, this direction may require continuous deployment at the inference stage, resulting in a significant order of inference overhead.

Another direction is removing the backdoor through channel suppression by evaluating spectral norm variation ~\cite{zheng2022data, ahmed2023ssda}. However,  since our attack relies on soft-prompts (a few hundred thousand parameters only) rather than weight channels,
removing the backdoor through parameter compression may significantly hamper the model's ability to generate reasonable task plans for benign input.

Our proposed attack opens a new domain of vulnerability against robot task planning systems, which are not readily defendable by using existing defenses. Hence, a formal defense investigation leveraging these existing defense directions is required, which is beyond the scope of our current work.